# Identifying emergency stages in Facebook posts of police departments with convolutional and recurrent neural networks and support vector machines

Nicolai Pogrebnyakov
Copenhagen Business School
Frederiksberg, Denmark
E-mail: nicolaip@cbs.dk

Edgar Maldonado
Metropolitan State University of Denver
Denver, CO, United States
E-mail: emaldon3@msudenver.edu

**Classification of social media posts in emergency response is an important practical problem: accurate classification can help automate processing of such messages and help other responders and the public react to emergencies in a timely fashion. This research focused on classifying Facebook messages of US police departments. Randomly selected 5,000 messages were used to train classifiers that distinguished between four categories of messages: emergency preparedness, response and recovery, as well as general engagement messages. Features were represented with bag-of-words and word2vec, and models were constructed using support vector machines (SVMs) and convolutional (CNNs) and recurrent neural networks (RNNs). The best performing classifier was an RNN with a custom-trained word2vec model to represent features, which achieved the F1 measure of 0.839.**

## 1. Introduction

The 2017 Atlantic Hurricane season was one of the most devastating on record for the Caribbean Islands and the US mainland, and also it has been the set of natural disasters where social media have played a significant role. Government and consulting firms are now using social media feeds to inform their decisions and provide a comprehensive picture of what is happening. The New York Times put it in simple words: "when disaster hits and landlines fail, social media is a lifeline" [1]. Social media seem to be able to deliver information to victims and responders at a faster pace than traditional emergency response protocols, so their use during emergencies is growing [2].

A commonly accepted distinction in emergency response is between three stages of an emergency: preparedness, response and post-emergency recovery [3]. During the preparedness stage, responders help the public understand the possible consequences of a disaster, prepare for them and draw contingency plans. The response phase involves dissemination of up-to-date information about the event, as well as possibly soliciting information from the public that could limit the negative consequences of the event (e.g., asking for help in locating a missing person). In the post-emergency recovery phase information about the end of the emergency or repairing any damage is disseminated [4].

Such stages-based view of emergencies can be naturally used in classifying social media messages that relate to a given emergency. Indeed, social media are increasingly being used at different stages of emergency response, from preparedness to actual disaster response [5-7]. They have become an important information sharing tool in emergencies of different scale, from large-scale disasters such as earthquakes [8] to smaller-scale emergency events, e.g. wildfires [9]. In turn, the public increasingly expects emergency responders to communicate through social media [10].

At the same time, these time-based stages are just one of the attributes that could be used in a scheme to categorize social messages. Researchers have also focused on a variety of other attributes of messages, sometimes specifically targeting the type of messages (e.g., monitoring fires [11]), and aspects of emergency response. Literature includes classifiers that analyze the credibility of information in messages [12], the location of message authors [13], or how informative a message is [14].

One of the primary emergency responders is police, who help mitigate the consequences of these events in addition to maintaining law and order on a daily basis. Like other responders, police increasingly rely on social media to disseminate information to the public [15]. And, similarly to other social media users, police provide updates not only during emergencies but also in "quiet times" [16]. In this paper, we focus on social messages from police departments because they have the most traffic on their Facebook accounts when compared with other emergency responders since they deal with all kind of emergencies, all year around.

Correctly identifying the topic of a message from an emergency responder, such as a police department, is an important problem. It can help the public better prepare themselves or react to an emergency. It may also help other responders (e.g., other police departments, firefighters and paramedics) coordinate their efforts better. For example, Latchal et al. [17] have suggested the development emergency-only hashtags or search terms to increase the efficiency of how information is disseminated during an emergency. Since whether a message relates to a future, current or past emergency is always relevant, classifying social media messages according to the three stages of the emergency should be part of most classification efforts. At the same time, manual categorization of messages by their authors—the police—according to a unified schema (e.g., with special-purpose tags for each category) is unlikely, as evidenced by the slow adoption of emergency communication standards, even within one country [18]. The results of automatic classification of emergency messages by the phase of emergency could be used standalone or as a first step for other classification schemes described above, such as the location of message authors or message credibility.

To that end, this research builds an automatic classifier of Facebook messages from police departments. The classifier distinguishes messages from the three emergency phases, as well as "engagement" messages posted by police departments to build trust with the community and provide updates on non-emergency related events. Classification is done with three methods: support vector machines, convolutional and recurrent neural networks. Different types of representing messages for classification were used, including bag-of-words and word2vec.

In the rest of the paper we briefly review prior work before discussing data and classification approaches, presenting results and providing concluding observations.

## 2. Literature Review

As more and more people and organizations take to social media to post and consume information, a major problem with social media use has been information overload. Using social media feeds seems a challenging task for an emergency response organization. Verma et al. [19] stated more than half of a decade ago how "(S)o much information is now broadcast during mass emergencies that it is infeasible for humans to effectively find it, much less organize, make sense of, and act on it". Given the amount of data that needs to be processed, computer methods have been used to assist in the categorization of social media emergency-related information [20].

Prior work has looked into the problem of classifying social media messages coming from emergency responders from several angles. Authors have used distinct dimensions: how relevant is the message [21], how likely a message is related to a specific situation [22]. Researchers furthermore have used different categories and focused on distinct aspects of information on social media. Imran et al. [23] developed one of the most exhaustive classifications of social media content in "high-impact events". Their study defined six categories of how social media messages can be classified: (1) by factual, subjective, or emotional content, (2) by information provided, (3) by information source, (4) by credibility, (5) by time, and (6) by location.

Using Imran's et al. categorization, we have developed a classification that applies to messages coming from emergency responders. With that goal, some of the categories in Imran's et al. study did not seem appropriate to use. Category 1 is not relevant for the discourse of an emergency responder. Categories 3 and 4 do not apply since the emergency responders already have credibility and the origin of its messages is understood. Finally, category 6 is also not relevant, since in most cases the specific police authority will have local or regional jurisdiction. Therefore we focus on category 5, classifying messages by time, or, in other words, by stages of emergency.

Our categorization of emergency messages can be used to triage messages by time. Therefore when messages have been received by the general public, they take appropriate action. For example, after a shooting the public needs to be sure that the emergency is over so they can go on with their daily activities, thus waiting for a message in the post-emergency category to arrive. Another example will be messages sent before an emergency requesting sandbags. Such messages will be categorized as emergency preparedness, and they will not be taken into account after the disaster.

Social media messages from emergency responders can also be classified using category 2, by information provided. Automatic categorization also can be used to assist responders when creating their own messages, so they can effectively convey needed information.

Several methods have been used to categorize social media messages, including Bayesian networks [24], clustering [25], neural networks [26], and support vector machines [27]. Our study uses support vector machines and two types of neural networks and compares their performance.

## 3. Data

Much of the prior work, both in the emergency realm and in more broad social media message classification tasks, has focused on classifying Twitter messages. We decided to use Facebook messages instead, and do so for two reasons: 1) while Twitter limits message length, Facebook posts can be longer, allowing for greater expressivity; 2) Facebook has greater reach: there are over 6 times as many worldwide users on Facebook as on Twitter as of August 2017 [28]. Thus Facebook represents a viable choice of platform.

We extracted 5,000 messages at random from Facebook posts by US police departments. To obtain the messages, we first collected a list of URLs of police departments from the websites usacops.com and policeapp.com. Links to the departments' Facebook profiles were obtained by a script that accessed each URL and checked for presence of a link to such profile. We then retrieved posts in the period from January 1, 2009 to November 1, 2016 for each profile, and selected a random set from the posts.

| Message Category | Description | Examples |
|---|---|---|
| Emergency preparedness | Preparedness information, tips on how to prepare for and behave in an emergency | "Every year we get reports of bears in the Bayport area. The mild weather has the bears already out and about throughout Minnesota. Time to bring out the bear-resistant garbage cans!"<br><br>"Pointers from the FTC about what to do if you suspect your computer has a virus or malware on it!"<br><br>"When driving/walking or cycling during the day, wear sunglasses to protect your eyes from glare." |
| Emergency Response | Update about an ongoing or past emergency | "Please avoid the area of SE 11th Drive and SE Wacahoota Road. A sinkhole has opened up in the area. Deputies have blocked passage on SE Wacahoota on either side of SE 11th Drive. Alachua County Public Works is responding to investigate."<br><br>"MISSING PERSON: SAPD Detectives request public's assistance in locating [person name] (62). Please contact the Santa Ana Police Watch Commander at [phone #] with any information regarding this missing person." |
| Post-Emergency and Recovery | Tips for mitigating effects of a past emergency, new information about a past emergency | "UPDATE: [Person name] has been found in good condition. Thank you to everyone for their assistance."<br><br>"TRAFFIC UPDATE: VDOT reporting Ivy Rd. is now open after extensive utility repairs yesterday & overnight. Safe travels on your morning commute!" |
| Engagement | Updates about the department's internal non-emergency operations, conversation with Facebook users | "17,000 Facebook Fans and Growing - Thank You Evesham Township! Spread the word to your friends and neighbors."<br><br>"ETPD would like to give a big THANK YOU to all the students and staff of Evans School. The school raised over $1200.00 to help support the ETPD K9 Unit." |

Table 1. Message Categories

That set was manually classified into one of four categories shown in Table 1. We recruited two raters on upwork.com, who were selected among five shortlisted candidates by completing a test assignment best (the assignment was a small sample of posts from the dataset). The raters were trained by explaining the classification schema and with a guided classification of a sample of posts. During actual classification, progress discussions with researchers were held after every 1,000 classified posts. The raters classified 80.1% posts similarly. The remainder was run through two rounds of classification on Amazon's Mechanical Turk, where the task was described using the schema and examples from Table 1. As a result, 94.8% of posts were classified similarly at least twice. Finally, the posts where there was still disagreement between the four ratings were classified by one of the researchers who cast the deciding vote.

## 4. Experiment Setup

### A. Constructing Features

We used several categories of features to represent the posts: as bag-of-words, parts of speech, descriptive features and word2vec.

*Bag-of-words (BOW)*. The sparse bag-of-words representation was used for SVM. The four message categories may differ in the topics they emphasize and, consequently, the vocabulary used within each category. We encoded posts as trigrams [29]. Extant studies have shown good performance of this social media message representation across a variety of domains, including sentiment analysis [30] and topic classification [31].

*Parts of speech (POS)*. Parts of speech features were included in SVM experiments to reflect possible differences in message style between categories. In particular, messages in the engagement category could be using fewer verbs than in the three emergency-related categories, as in these examples:

*"For all those inquiring about Halloween Trick or Treating, it will be on Saturday, October 26, 2013 from 6pm-9pm! Happy Haunting!"*

*"Even K-9 Dooly needs a little play time! He's not only a hard worker, but a great companion to this pup!"*

Additionally, we conjecture that the three emergency-related categories lend itself to the use of verbs in specific tenses. Preparedness messages may use proportionally more verbs in the future tense:

*"The Sam Varnedoe drawbridge will LIKELY remain CLOSED until tomorrow morning"*

Emergency response messages may employ more present-tense verbs:

*"Power outages again being reported South and Southeast of Greenville"*

Recovery messages may contain a mix of past, present and future, referring to a past emergency (past) and suggesting recovery actions (present and future):

*"UPDATE: New Jersey American Water has completed repairs to the water main in Manville. All water service is now back to normal."*

*Descriptive features (DESC)*. We included the following 5 features that further characterize a message and are not included in the BOW or POS features: message length (the number of words), the number of likes, the number of exclamation and question marks in the message and the proportion of capital letters in the text.

*B. Word2vec Embeddings*

While the BOW feature representation identifies words that distinguish between documents in a corpus well and is relatively simple, it is not an optimal choice for some classifiers. First, BOW results in sparse feature vectors as well as a loss of semantic information (e.g., synonymic similarities of words), which hinders filter learning in CNNs [32]. Second, it discards word order, which is important in training RNNs (see section 4.F). To overcome these limitations, feature representations based on word distributions within text have been proposed, such as latent Dirichlet allocation (LDA) [33] and word2vec [34]. We thus also represented messages with word2vec embeddings, and used exclusively them in the neural network models.

In word2vec, a model is trained on a large text corpus that represents each word as a vector of predefined dimensionality. The resulting vectors contain semantic, syntactic and morphological properties of words, such as singular/plural noun forms or verb tenses. These properties are not manually specified in the model: instead, they are learned from raw text [35].

Two word2vec embeddings were used. In one, features were represented with publicly available word2vec embeddings trained on 100 billion words from the Google News dataset [36] (the *"generic"* feature set, labeled GW2V). In the other (the *"custom"* feature set, labeled CW2V), embeddings were trained on all 1,011,144 police Facebook messages over the time period under consideration from which the 5,000 messages in our dataset had been randomly selected. GW2V and CW2V were used standalone and in conjunction with DESC features.

In the GW2V and CW2V representations the maximum length of messages was set to 100 words. 84.4% of messages in the dataset were shorter than that, and were padded with zero vectors to that length. Longer messages were truncated to the first 100 words.

All experiments were run with word2vec embeddings and the DESC representation. In addition, SVM experiments were run with BOW and POS features.

*C. Data Transformation*

Each message was tokenized and lemmatized. We converted the names of individual days of the week into one term, [DayOfWeek], and similarly names of months into [Month]. URLs, dates, phone numbers, all other numbers, email addresses and Twitter-style "@username" handles were each replaced by a respective generic term (e.g., [URL]). Parts of speech were tagged with Penn Treebank tags using the NLTK tagger.

| Set | Prepared-ness | Response | Post-Emergency | Engage-ment | Total |
|---|---|---|---|---|---|
| Training | 573 | 364 | 931 | 1,627 | 3,495 |
| Test | 266 | 139 | 389 | 711 | 1,505 |

Table 2. The Number of Examples in the Training and Test Sets by Message Classes

| Measure | Min | Median | Mean | Max | St. dev. |
|---|---|---|---|---|---|
| Words in message | 0 | 25 | 62.8 | 2,847 | 126.5 |
| Message likes | 0 | 12 | 95.5 | 118,201 | 1,704 |

Table 3. Descriptive Statistics of Messages in Dataset

With BOW, features were represented in three ways. In the "Bool" representation, only the presence or absence of a feature was recorded (0/1). "Freq" included the number of occurrences of each term or feature in the message. With the "Tfidf" representation, tf-idf weighting of features was performed.

We converted the manually classified messages in the corpus into feature representations described above and split them into a training and test datasets (Table 2 shows the number of examples in each set).

Because the dataset is unbalanced, with considerable variation in the number of posts across categories, we used the F1 measure to evaluate the algorithm performance instead of accuracy.

Table 3 shows descriptive statistics of the dataset used in classification.

*D. Support Vector Machines*

SVM has been extensively used in classifying text [37, 38]. Several features of SVM make it particularly suitable for this task. Compared to some other classification methods such as logistic regression, there is less need for exhaustive feature selection, and SVM is not as sensitive to outliers [37]. Additionally, it performs well on sparse data, which is characteristic of text documents represented by feature vectors [39].

SVM's goal is to separate data points of two classes with a boundary that is located within the widest possible margin between these classes. The placement of the boundary separating the classes is heavily influenced by support vectors, which are training data points that lie closest to the boundary [40]. Different kernel functions control the complexity of the boundary, with commonly used kernels being linear, polynomial and Gaussian.

In this study an SVM with the linear kernel was trained using L2 regularization. The value of the hyperparameter $C$ was first selected using 10-fold cross-validation, which yielded $C = 0.001$.

*E. Convolutional Neural Networks*

Convolutional neural networks (CNNs) have found widespread use in computer vision [41, 42], and have been applied to text categorization as well [26, 43]. They have fewer parameters and train faster than regular feed-forward neural networks with a similar number and size of layers [42]. This is achieved with convolutional and pooling layers.

In convolutional layers each unit is connected only to a small number of units in the preceding layer (its receptive field). These layers contain learnable kernels, or filters, each of which detects a particular feature of a given size at each location in the input [41]. Individual kernels are arranged in two-dimensional layers (feature maps), which in turn are stacked across the third dimension. With multiple convolution layers, the network learns increasingly complex features, e.g. from edges of objects in lower layers to dog faces in upper layers in image recognition [44].

Pooling layers are another commonly used type of layer in CNNs. They apply an aggregation operation, such as average or max, to a group of neighboring units in the same feature map from the preceding layer [42, 45]. This reduces the dimensionality of hidden representation passing through the network and helps speed up training.

A potential problem with deep neural networks is overfitting. Several methods have been proposed to address it, including adding regularization to the cost function or applying dropout, where units are randomly removed from the network during training [45, 46].

The following CNN architecture was used in our study:

- Input layer
- Convolutional layer with 100 feature maps, rectangular kernels of size $5 \times 1$ and stride $2 \times 1$
- Max-pooling layer with kernel size $5 \times 1$
- Convolutional layer with 200 feature maps, kernel size $3 \times 1$ and stride $2 \times 1$
- Max-pooling layer with kernel size $3 \times 1$
- Dropout at 0.5 rate
- Fully connected layer with 200 units
- Output softmax layer

Learning rate was 0.001, the convolutional and fully connected layers used ReLU activation, minibatch size of 50 was used, and batch normalization was performed. We did not use early stopping. All these hyperparameter values were selected using the random search strategy [47].

In particular, rectangular-shaped filters (with greater height than width) were found to perform better than commonly used square-shaped. E.g., in the input layer filters span 5 words and 1 word2vec feature. Strides were also made longer vertically than vertically. Their better performance might be explained by an analogy with *n*-grams, which perform better with *n* > 1. In our case, filters for one feature of several input words at a time may work better than for several features and several words (square-shaped) because neighboring words exhibit greater regularities than multiple neighboring features of neighboring words, although this requires additional research.

The size of the input and hidden layers varied with feature representation: e.g. in the GW2V+DESC representation the input vectors are (300 generic word2vec features + 5 DESC features) × 100 words per message = 30,500 in length, which is the number of units in the input layer.

*F. Recurrent Neural Networks*

In contrast to SVM and feed-forward neural networks such as CNNs, which process entire input vectors, recurrent neural networks (RNNs) process inputs sequentially. This makes them well suited for working with data that can be separated into blocks, e.g., words of a sentence, and they have been successfully used in natural language processing tasks such as text classification [48, 49] and machine translation [50].

RNNs take as input a sequence of vectors $\mathbf{x} = (\mathbf{x}_0, \ldots, \mathbf{x}_{T-1})$, compute a sequence of hidden-state vectors $\mathbf{h} = (\mathbf{h}_0, \ldots, \mathbf{h}_{T-1})$ and output a sequence $\mathbf{y} = (\mathbf{y}_0, \ldots, \mathbf{y}_{T-1})$ [51]. Inputs are processed in sequence, and the output of an individual step in the sequence $\mathbf{y}_i$ depends not only on the corresponding input vector $\mathbf{x}_i$, but also on the hidden state at the previous step $\mathbf{h}_{i-1}$ [52]. The hidden state from the previous step allows the network to use dependencies among input vectors, e.g. between neighboring words in the input sentence. During training, the network is "unrolled" for the specified number of sequence steps, which allows forward and backward propagation through time [53].

One drawback of such RNN architecture is over-reliance on data immediately preceding a given input vector $\mathbf{x}_i$, as the influence of inputs $(\mathbf{x}_0, \ldots, \mathbf{x}_k)$, $k < i$ earlier in the sequence either grows exponentially or diminishes to close to zero (the exploding/vanishing gradients problem) [53, 54]. To introduce longer-term dependencies and combat the problem of vanishing or exploding gradients when training networks with multiple layers or time steps, unit architectures with gates have been introduced: Long Short-Term Memory (LSTM) [55] and the Gated Recurrent Unit (GRU) [56].

We used a relatively simple network architecture of a single GRU cell with ReLU activation. The cell was unrolled over 100 steps, with a softmax output layer.

## 5. Results

*A. Support Vector Machines*

*1) BOW Feature Set*

The results are shown in Table 4. The Tfidf feature representation resulted in best performance, with F1 measure of 0.810 on the test set.

| Represe-ntation | F Measure by Class[a] | | | | $F_{avg}$ |
|---|---|---|---|---|---|
| | $F_{prep}$ | $F_{resp}$ | $F_{post}$ | $F_{eng}$ | |
| Bool | 0.429 | 0.674 | 0.709 | 0.802 | 0.697 |
| Freq | 0.454 | 0.678 | 0.698 | 0.793 | 0.696 |
| Tfidf | 0.668 | 0.777 | 0.787 | 0.886 | **0.810** |

[a.] Using binary (Bool), frequency-based (Freq) and tfidf-normalized (Tfidf) feature representation. Results for this and the other SVM experiments are for the test set.

Table 4. Results of SVM on the BOW-only Feature Set

| Preparedness | Response | Post-Emergency | Engagement |
|---|---|---|---|
| [URL] | investigation | arrest | happy |
| weather | info | update | event |
| your | watch log | was | community |
| tip | burglary | injury | national |
| warning | robbery | suspect | officer |
| closure | [number] [URL] | wreck | our |
| scam | call | juvenile | program |
| travel | scene | vehicle wreck | activity [URL] |
| storm | on scene | has | great |
| drive | sought | arrest [URL] | congratulation |

Table 5. Most Important Features for Each Class on the BOW Feature Set

We examined the most important features for each class from SVM classifier's output (see Table 5). The Preparedness category includes terms such as "warning", "reminder", "tip" that relay information about an upcoming event while encouraging the reader to remain attentive to that information. Many posts in this category include weather warnings, which is reflected in weather-related terms: "weather", "snow". "Closure" in that category refers to upcoming road closures.

In the Response category, terms describing specific crimes ("burglary", "robbery") are among the most important. Also important are terms that encourage the public to provide information ("call", "sought"), as in these examples:

*A male suspect is being sought in connection with a strong armed robbery that occurred at a S.R. 16/I-95 motel (Best Western) room Saturday afternoon.*

*Information sought in assault offense - 3500 Gaston Avenue*

In the Post-Emergency and Recovery category, words referring to a past event are among the most important: "was", "arrest" (which in the context of police messages indicates a completion of e.g. a manhunt), and "has" (the auxiliary "has" of the present perfect tense, as in "has completed"). Examples:

*The demonstration at Saint Paul and Lexington Street has disbanded. No traffic was impacted.*

*Gun Arrest: 200 block N. Howard Street. Central District midnight shift officers arrest a 20-year old man & recover a loaded handgun.*

The terms "injury", "suspect", "wreck" and "juvenile" provide additional details about the past event, as in this example:

*Clearwater Police and Clearwater Fire & Rescue responded to a car fire shortly after 4 p.m. on Memorial Causeway. An elderly couple was on the way to the beach when someone in the car behind them noticed flames near the bottom of the car and alerted them. The occupants were able to escape without injury.*

Finally, the top words in the Engagement category include those promoting positive sentiment ("happy", "great"), providing internal information about the department ("our", "office") and engaging users ("congratulation"):

*This week is National Public Safety Telecommunicator's Week and we are proud to celebrate the job that our Central Communications staff perform daily!*

*Congratulations to the wonderful, strong and determined individuals who graduated today from the Women Against Crime program. Thank you Sgt. Hoffman for your dedication to our community.*

*2) POS + DESC Feature Set*

Performance on this feature set was markedly weaker than on BOW (see Table 6). The best-performing feature representation was Tfidf, however it only achieved F1 measure value of 0.487.

| Represe-ntation | F Measure by Class | | | | $F_{avg}$ |
|---|---|---|---|---|---|
| | $F_{prep}$ | $F_{resp}$ | $F_{post}$ | $F_{eng}$ | |
| Bool | 0.023 | 0.049 | 0.443 | 0.683 | 0.425 |
| Freq | 0.278 | 0.084 | 0.461 | 0.687 | 0.482 |
| Tfidf | 0.272 | 0.083 | 0.477 | 0.693 | **0.487** |

Table 6. Results of SVM on the POS + DESC Feature Set

| Preparedness | Response | Post-Emergency | Engagement |
|---|---|---|---|
| IN | JJ | VBN | WP |
| CC | DT | VBD | VerbFuture |
| VBG | NN | NN | JJR |
| JJ | VBG | CD | PDT |
| RB | IN | IN | TO |
| VerbPresent | NNP | RB | NNS |
| NN | VerbPresent | NNP | VerbPast |
| DT | VBN | JJ | PRP |
| WRB | CC | NNS | RBS |
| NNP | MD | DT | WDT |

*Note:* CC coordinating conjunction: "and", "or"; CD cardinal number "two"; DT determiner: "all"; IN subordinating conjunction: "though"; JJ adjective: "high"; JJR comparative adjective: "higher"; MD modal verb "can", as in "can do"; NN singular noun: "tree"; NNP proper noun "Colorado"; NNS plural noun "trees"; PDT predeterminer: "all"; POS possessive ending: "'s", as in "Mike's"; PRP personal pronoun "her"; RB adverb: "highly"; RBS superlative adverb: "most", as in "most interesting"; TO the word "to"; VB infinitive verb: "ride"; VBD past verb: "rode"; VBG gerund: "riding"; VBN past participle: "ridden"; VerbPast past-tense verb clause: "has done"; VerbPresent Present-tense verb clause: "is doing"; VerbFuture future-tense verb clause: "will be doing"; WDT wh-determiner: "whose"; WP wh-pronoun: "who"; WRB wh-adverb: "where"

Table 7. Most Important Features for Each Class on the POS + DESC Feature Set

From Table 7 it is apparent why the classifier performed so poorly with this feature set, particularly on the Response and Preparedness classes where the best classifier achieved only 0.083 and 0.272 accuracy for each class. The top most important features for these two classes are essentially the same: the only feature from the Preparedness class that is not also in Response is RB (but it is among the top features in the Post-Emergency and Recovery class). The top features in the Post-Emergency class are more different from the Preparedness and Response classes, but there are still significant overlaps.

Thus our conjecture that messages in each of the three emergency stages might predominantly use verbs in a specific tense was not supported.

By contrast with the three emergency-related classes, most top features in the Engagement class are unique to that class, which explains the relatively decent performance of the classifier on that class. The top features are all POS tags, rather than descriptive features.

DESC features did not have a substantial influence on classification. The number of likes is the $25^{th}$, and the proportion of capital letters the $26^{th}$ most important feature in the Engagement class. The other descriptive features were not important in classifying the dataset. Consider the following message from the Response category, where each word was tagged with a POS tag (see the footnote above for explanations of tags). Note that it contains most of the top features from all three emergency-related classes:

*A   white male was dropped off by a   person driving a*
 DT  JJ   NN  VBD  VBN   RP IN DT   NN   VBG  DT
*white 4 door sedan in front of Wal-Mart, 5600 N Henry*
  JJ  #  NN   NN  IN NN  IN    NNP    #  NNP  NNP
*Blvd, Stockbridge, Ga. The white male shoplifted a*
 NNP    NNP      NNP DT  JJ   NN    VBD   DT
*Samsung 32 inch flat screen TV and left through a fire exit*
  NNP   #  NN  JJ   JJ    NN CC VBD  RP   DT  NN NN
*to the same white 4 door sedan waiting to pick him up.*
 TO DT  JJ    JJ  #  NN   NN    VBG  TO  VB PRP RP
*Anyone with information identifying the suspects, please*
  NN   IN     NN         VBG    DT  NNS      VBP
*contact Detective Baugh.*
  JJ       NNP    NNP

| Represe-ntation | *F* Measure by Class | | | | $F_{avg}$ |
|---|---|---|---|---|---|
| | $F_{prep}$ | $F_{resp}$ | $F_{post}$ | $F_{eng}$ | |
| Bool | 0.430 | 0.697 | 0.728 | 0.807 | 0.707 |
| Freq | 0.499 | 0.713 | 0.725 | 0.817 | 0.726 |
| Tfidf | 0.682 | 0.791 | 0.783 | 0.884 | **0.813** |

Table 8. Results of SVM on the Combined BOW + POS + DESC Feature Set

| Representation | *F* Measure by Class | | | | $F_{avg}$ |
|---|---|---|---|---|---|
| | $F_{prep}$ | $F_{resp}$ | $F_{post}$ | $F_{eng}$ | |
| GW2V | 0.579 | 0.547 | 0.738 | 0.819 | 0.716 |
| GW2V+DESC | 0.598 | 0.529 | 0.745 | 0.816 | 0.717 |
| CW2V | 0.677 | 0.605 | 0.688 | 0.852 | 0.745 |
| CW2V+DESC | 0.675 | 0.610 | 0.693 | 0.857 | **0.748** |

Table 9. Results of SVM on the Generic and Custom word2vec Sets

(Note that the tagger labeled several words incorrectly, such as "door", which is an adjective here, "contact", which is a verb and "Detective", which is a noun.)

*3) Combined BOW + POS + DESC Set*
Combining the two sets results in improved performance across all representations compared to the BOW only set (see Table 8).

The Freq representation experienced the highest gain (0.030), while in the Tfidf representation performance improved to 0.813, which is the highest score for the SVM classifier across all feature sets.

*4) Word2vec Set*
Table 9 shows the results on the generic (GW2V) and custom (CW2V) word2vec sets.

This feature set showed lower performance than BOW sets, yet greater than the POS + DESC set. Custom word2vec representations performed better than generic ones, both with DESC features and without them, and the addition of DESC features improved on the results of word2vec-only features.

*5) Additional SVM Experiments*
In addition to experiments reported above, we performed several experiments that did not result in improved performance.

*Ngrams*. While the above experiments were conducted with trigram BOW, in this experiment we used unigram and bigram BOW features instead.

*Two-Letter POS*. We also used two-letter POS tags instead of three-letter ones. The third letter in the tags is used to provide additional information about a specific part of speech: e.g. NN is a singular noun while NNS is a plural noun. This experiment sought to check whether less specific two-letter tags would result in better performance.

*Synonyms*. With this approach, we replaced groups of words in the message with their synonyms, which resulted in a reduced dimensionality of the feature space. We ranked all words extracted from the message corpus by tf-idf and, starting from the top, replaced a particular word with its synonym if the synonym had a higher tf-idf score. The list of synonyms from WordNet was used, and part of speech was included in synonym lookup.

The results using the Tfidf feature representation are shown in Table 10. Although the results vary, the best performing classifiers are the 2-letter POS and bigrams. At 0.812 accuracy, their performance is only slightly below that of the combined dataset of trigram BOW + full POS + DESC. This indicates that the original choice of full POS tags works better than tags shortened to 2 letters, and that trigrams result in better performance than bigrams, but also that unigrams and bigrams, which require less computational resources, result in performance comparable to trigrams.

### B. Generic and Custom Word2vec Representations

Training a custom word2vec model increased classification performance with all classifiers, even though we opted for smaller dimensionality of feature vectors in the custom representation than in the generic one (100 vs. 300 features respectively). The gains ranged from 1.2% for RNN on word2vec-only features to 6.4% for CNN on word2vec-only features.

We examined words with the most similar vectors in the generic and custom word2vec set, as well as WordNet synonyms, for top words for each class from Table 5. An example for the word "weather", which is the top word for the Preparedness class, is shown in Table 11.

Although the number of similar words and their overlap varies for different words, custom vectors, unsurprisingly, capture more of the domain specifics. In this example for "weather", custom embeddings contain different types of weather events. Generic ones also include weather events, however, many of them are formed by adding "weather" to a modifier: e.g., "wet weather" instead of "rain". And the number of WordNet synonyms is limited, in addition to them not including a variety of weather conditions, which helps explain poor performance of the "Synonyms" experiment in SVM. In other instances (e.g., for "investigation", not shown here), generic embeddings contain various misspellings of the word while custom ones provide synonyms ("inquiry") and WordNet provides on the order of 20 synonyms.

These results suggest that gains in classification performance stem from the custom word2vec representation's better ability of capturing domain-specific similarities between words. Thus, accuracy in the supervised classification task can be increased by unsupervised methods (here, training a domain-specific word2vec model).

### C. Convolutional Neural Networks

Table 12 shows CNN results. The metrics were calculated on the test set after 10-fold cross-validation.

Custom word2vec embeddings with descriptive features showed the best performance for this classifier, although at 0.784 it is lower than SVM or RNN results.

### D. Recurrent Neural Networks

The results are shown in Table 13. Similarly to CNN, CW2V + DESC performed best, and the use of this representation with RNN resulted in the best F1 measure, 0.839, across all classifiers.

### E. Summary Observations

To sum up the results of our experiments: of the classifiers, RNN performed best, followed by SVM and CNN. In feature representations, custom word2vec embeddings with descriptive features showed the best performance, followed by tf-idf weighted bag-of-words.

Among the four message classes, the Engagement category has consistently had the highest F1 score across all classifiers and feature representations. Of the emergency- related

| Feature set[b] | *F* Measure by Class | | | | $F_{avg}$ |
|---|---|---|---|---|---|
| | $F_{prep}$ | $F_{resp}$ | $F_{post}$ | $F_{eng}$ | |
| Unigrams | 0.682 | 0.759 | 0.787 | 0.887 | 0.811 |
| Bigrams | 0.664 | 0.801 | 0.788 | 0.883 | **0.812** |
| 2-letter POS | 0.683 | 0.778 | 0.779 | 0.887 | **0.812** |
| Synonyms | 0.633 | 0.751 | 0.755 | 0.870 | 0.787 |

[b]. All feature sets include BOW, POS and DESC features. BOW are trigrams except in Unigrams and Bigrams.

Table 10. Results of Additional SVM Experiments

| **Generic Word2vec** | wet weather, inclement weather, wintry weather, wintery weather, weatherwise, stormy weather, unusually mild, unseasonably warm, wintry conditions |
|---|---|
| **Custom Word2vec** | storm, drought, blizzard, snowstorm, forecast, NWS[c], storms, thunderstorm, heat, hurricane |
| **WordNet Synonyms** | weather condition, conditions, atmospheric condition |

[c]. NWS stands for the US National Weather Service

Table 11. Words with the Most Similar Word2vec Vectors and WordNet Synonyms for the Word "Weather"

| **Representation** | *F* Measure by Class[d] | | | | $F_{avg}$ |
|---|---|---|---|---|---|
| | $F_{prep}$ | $F_{resp}$ | $F_{post}$ | $F_{eng}$ | |
| GW2V | 0.643 | 0.642 | 0.745 | 0.841 | 0.752 |
| GW2V+DESC | 0.683 | 0.618 | 0.756 | 0.871 | 0.771 |
| CW2V | 0.734 | 0.677 | 0.757 | 0.846 | 0.780 |
| CW2V+DESC | 0.678 | 0.698 | 0.773 | 0.863 | **0.784** |

[d]. In CNN and RNN results GW2V refers to generic, CW2V to custom W2V feature representation.

Table 12. Convolutional Neural Network Results

| **Representation** | *F* Measure by Class | | | | $F_{avg}$ |
|---|---|---|---|---|---|
| | $F_{prep}$ | $F_{resp}$ | $F_{post}$ | $F_{eng}$ | |
| GW2V | 0.740 | 0.720 | 0.780 | 0.886 | 0.810 |
| GW2V+DESC | 0.711 | 0.699 | 0.808 | 0.898 | 0.813 |
| CW2V | 0.735 | 0.725 | 0.810 | 0.894 | 0.820 |
| CW2V+DESC | 0.776 | 0.785 | 0.806 | 0.902 | **0.839** |

Table 13. Recurrent Neural Network Results

categories, Post-Emergency and Recovery has been the most correctly classified.

## 6. Discussion and Conclusion

In this research, we performed classification of Facebook messages by police departments using four classes: three emergency-related classes, one for each stage of an emergency (Preparedness, Response and Post-Emergency and Recovery), as well as the Engagement class that reflected conversations with Facebook users and community engagement. The best classifier (RNN) achieved accuracy of 0.839 on the test set. The recurrent neural network classifier also outperformed SVM, which in turn performed better than the convolutional neural network.

Since information overload seems to be one of the main barriers for the adoption of social media by emergency managers, automatic classification of messages is a reasonable approach to increasing the capability of emergency responders to process and accurately filter information. We see two potential uses of such classifier: by other emergency responders and by the general public. Using classifiers such as the one introduced in this study, emergency responders can initiate the triage of critical messages depending on the emergency phase they pertain to, as well as improve coordination between emergency responders during emergency preparedness, response and post-emergency recovery. The classifier may complement other, dedicated channels of communication between police, firefighters and paramedics to increase each other's situational awareness.

The results of this research may also be used to help channel the public's attention during different phases of an emergency. As a specific example, the classifier can be built into a Facebook app that monitors local police department Facebook accounts and alerts the user with only those messages that the user deems most pressing, for example ones categorized as Response or Post-Emergency and Recovery. This may help increase user attention to the types of messages or alerts that the user prefers to receive.

Future research can use a similar methodology to study social media messages from other emergency responders to look for similarities and differences. Classifying messages on dimensions other than emergency stage can also be performed, for example focusing on the type of information provided (e.g., property damage or missing person alerts).